\documentclass[conference,10pt,a4paper,final]{IEEEtran}

\newlength{\flexwidth}
\usepackage{amsmath,amssymb,amsthm,fixmath}
\usepackage{mathtools}
\usepackage{accents}
\usepackage{color,colortbl}
\usepackage{xcolor}
\usepackage{siunitx}
\usepackage{cuted}
\usepackage{multirow,booktabs}
\usepackage{subcaption}
\usepackage[inline]{enumitem}
\usepackage{optidef}
\usepackage{graphicx}
\usepackage{lipsum}
\DeclareMathOperator{\med}{med}
\DeclareMathOperator{\std}{std}

\usepackage{epstopdf}
\usepackage[textsize=tiny,colorinlistoftodos]{todonotes}
\makeatletter
\define@key{todonotes}{bh}[]{
	\setkeys{todonotes}{author=\textbf{Bin}, color=lime!30}}%
\define@key{todonotes}{zf}[]{
	\setkeys{todonotes}{author=\textbf{Zexin}, color=blue!30}}%

\makeatother
\newif\ifshownotes
\shownotesfalse
\usepackage[normalem]{ulem}
\newcommand{\revise}[2]{{\color{red}\sout{#1}}{\color{blue}#2}}
\renewcommand{\revise}[2]{#2} 

\usepackage[shortcuts,acronym,automake]{glossaries}
\makeglossaries
\newacronym{ue}{UE}{User Equipment}
\newacronym{bs}{BS}{base station}
\newacronym{csi}{CSI}{Channel state information}
\newacronym{cnn}{CNN}{Convolutional Neural Network}
\newacronym{fl}{FL}{Federated learning}
\newacronym{iot}{IoT}{Internet of Things}
\newacronym{mmwave}{mmWave}{millimeter-wave}
\newacronym{b5g}{B5G}{Beyond-Fifth-Generation}
\newacronym{6g}{6G}{Sixth Generation}
\newacronym{ml}{ML}{Machine learning}
\newacronym{sbs}{SBS}{small base station}
\newacronym{mu}{MU}{mobile user}
\newacronym{mbs}{MBS}{macro base station}
\newacronym{mse}{MSE}{Mean Squared Error}
\newacronym{cl}{CL}{centralized learning}
\newacronym{uav}{UAV}{unmanned aerial vehicle}
\newacronym{bme}{BME}{Bayesian Model Ensemble}
\newacronym{iid}{IID}{independent and identically distributed}
\newacronym{raf}{RAF}{robust aggregation function}
\newacronym{sgd}{SGD}{stochastic gradient descend}
\newacronym{cdf}{CDF}{cumulative distribution function}
\newacronym{lid}{LID}{local intrinsic dimensionality}
\newacronym{llpf}{LLPF}{local loss pre-filtering}
\newacronym{mitm}{MITM}{man-in-the-middle}
\newacronym{ae}{AE}{adversary entitie}

\usepackage{tcolorbox}
\tcbuselibrary{many}

\newtcbtheorem[]{alg}{Algorithm}{fonttitle=\bfseries}{alg}

\usepackage[norelsize,linesnumbered,ruled]{algorithm2e}
\SetKwRepeat{Do}{do}{while}
\makeatletter
\newcommand{\removelatexerror} {\let\@latex@error\@gobble}
\makeatother

\begin{document}
	
	\title{{Robust Federated Learning for Wireless Networks:\\A Demonstration with Channel Estimation}}
	
	\author{
		\IEEEauthorblockN{Zexin~Fang\IEEEauthorrefmark{1},~Bin~Han\IEEEauthorrefmark{1},~and~Hans~D.~Schotten\IEEEauthorrefmark{1}\IEEEauthorrefmark{2}}
		\IEEEauthorblockA{
  \IEEEauthorrefmark{1}{University of Kaiserslautern (RPTU), Germany}\\\IEEEauthorrefmark{2}{German Research Center for Artificial Intelligence (DFKI), Germany}
		}
	}
	
	\bstctlcite{IEEEexample:BSTcontrol}
	
	\maketitle

	\begin{abstract}
		\gls{fl} offers a privacy-preserving collaborative approach for training models in wireless networks, with channel estimation emerging as a promising application. Despite extensive studies on FL-{empowered} channel estimation, the security concerns associated with \gls{fl} require meticulous attention. In a scenario where \glspl{sbs} serve as local models trained on cached data, and a \gls{mbs} functions as the global model setting, an attacker can exploit the vulnerability of \gls{fl}, launching attacks with various adversarial attacks or deployment tactics. In this paper, we analyze such vulnerabilities, corresponding solutions were brought forth, and validated through simulation.
	\end{abstract}
	
	
	\IEEEpeerreviewmaketitle
	
	\glsresetall

	\section{Introduction}\label{sec:introduction}

     \gls{fl} empowers devices to collectively train a global model without sharing raw data, a crucial aspect for maintaining user data confidentiality in wireless networks \cite{wirelessFL}.  On-device {computing} units, such as TrueNorth and Snapdragon neural processors, further enable {\gls{fl}} with the necessary hardware solutions \cite{WirelessSUV}. Retaining data at its origin and employing on-device learning enable swift responses to real-time events in latency-sensitive applications. \gls{fl} is recognized as a promising enhancement for wireless networks in various aspects, such as resource allocation \cite{VLCFL, ResourceFL} as well as beamforming design \cite{elbir2020federated}. On the other hand, the global shift towards higher frequencies, attributed to their abundant spectrum resources, will raise new challenges. For instance
     , compared to \revise{cellular signals}{sub-6GHz systems}, \gls{mmwave} \revise{signals}{systems} face a more complex environment, characterized by higher scattering, penetration losses, and path loss for fixed transmitter and receiver gain \cite{mmWFL}. Especially, due to significantly shorter channel coherent time, conventional channel estimation techniques become impractical. \gls{ml} based channel estimation techniques have demonstrated their effectiveness in improving estimation performance, as well as reducing the estimation time \cite{mmWave6G}. 

     Given the inherent high dynamics of high-frequency radio channels, which can exhibit diverse characteristics across varying weather conditions, timely model updates based on real-time data are crucial. With \gls{fl}, the model can undergo rapid updates without substantial communication overhead. Meanwhile, the training will be carried out parallel, further reducing the update time. While compelling, many recent studies involve training local models with edge devices, followed by the base station receiving and aggregating model updates to update the global model \cite{mmWFL,FLDeviceJeon,FLDeviceAmiri}. On-device training reduces communication overhead, yet synchronizing diverse edge devices on large-scale, particularly on resource-constrained \gls{mu}, poses a significant challenge. Thus, effort have been made towards \gls{sbs} level training, with the \glspl{sbs} training on a small dataset cached from nearby \glspl{mu}, and the model aggregation is performed on a \gls{mbs} \cite{FLBaseAbad,FLBaseHou}. This \gls{fl} strategy aims to provide more reliable and timely model updates. Most recently, research focus have been shifted to scenarios where both \gls{bs} and edge devices are capable  of training local models, and the aggregation is performed on the \gls{bs} \cite{FLBaseHong,FLBaseCao}.  
     

     Threats of \gls{fl}-based channel estimation can be categorized in \revise{2}{two} aspects: \begin{enumerate*}[label=\emph{\roman*)}]
        \item malicious local models providing malicious model updates
        \item data from poisoned sources. 
    \end{enumerate*}
     In \gls{bs}-based training, compromising a \gls{bs} \revise{to be}{into} a malicious local model is unlikely, yet the security concern persists with the vulnerability to poisoned data sources. However, there is still overlap between security concern of those frameworks. In instances of data poisoning, an attacker may deploy a substantial set of malicious \gls{mu} to execute data poisoning attacks on a specific \gls{bs}. In such scenarios, the impact is akin to involving a malicious local model. 

     In this work we consider data poisoning attacks on \gls{sbs}. The attacks may launch widely, or target in one specific \gls{sbs}, mirroring scenarios involving malicious local models. Firstly, we evaluate the vulnerability to adversarial attacks on channel estimation models. Secondly, we propose novel aggregation functions leveraging \gls{bme}, which ensures resilience against attacks without compromising estimation performance. Thirdly, we employ a pre-filtering strategy based on local loss distribution to mitigate attacks that are widely launched. Our simulation results highlight the effectiveness of our methods in mitigating diverse attacks, surpassing conventional approaches, and offering insights to bolster the security and resilience of \gls{fl} in wireless communication.
     
	\section{Preliminaries}\label{sec:problem_setup}
    \subsection{Channel estimation model}
    \gls{ml}-based models are trained with algorithms to learn underlying patterns and correlations in data, subsequently enabling them to predict patterns in new data. The channel estimation model $h$ can be described as follows,
    \begin{equation}
    \mathbf{y} = h(\mathbf{x},\mathbf{w}): \mathbb{P}^{m\times n} \rightarrow \mathbb{C}^{m\times n},
    \end{equation}
    where $\mathbf{x}$ represents the input pilot signal, $\mathbf{x}$ takes the form of a matrix $\mathbb{P}^{m\times n}$. The output $\mathbf{y}$ signifies the estimation result of the channel estimation model, with $\mathbb{C}^{m\times n}$ denoting the predicted \gls{csi} matrix. The parameter $\mathbf{w}$ corresponds to the weights of the channel estimation model. 
    
    By integrating \gls{fl}, we can refine the model by fine-tuning it using cached datasets, denoted as $\mathcal{B} = \{\mathcal{B}_{n}^{t} : 1 \leq n \leq N,  1 \leq t \leq T \}$, where $n$ represents the number of \gls{sbs} and $t$ denotes the federation round. At each federation round, local models undergo $K$ steps of \gls{sgd}, defined by the update rule:  
    \begin{equation}
    \mathbf{w}_n \leftarrow \mathbf{w}_n - \eta_k \nabla \ell(\mathcal{B}_k, \mathbf{w}_n), 
    \end{equation}
    where $\mathbf{w}_n$ is the model weights for each local model, and $\ell$ is the loss function. $\eta_k$ and $\mathcal{B}_k$ denote the step size and mini-batch at $\mathrm{k}\textsuperscript{th}$ step, respectively, with $\mathcal{B}_k\in \mathcal{B}_n^t$.
    \subsection{Robust model aggregation}\label{subsec:rma}
    Following the fine-tuning of the local model, the weight updates, denoted as $\mathcal{W} = \{\mathbf{w}_{n}^{t} : 1 \leq n \leq N,  1 \leq t \leq T \}$, are aggregated to create a new global model with weight $\mathbf{w}_g^t$. The standard aggregation function, \emph{FedAvg}, can be described as follows:
    \begin{equation}
    \mathbf{w}_g^t = \sum\limits^N_{n=1}\frac{l_n}{\sum(\mathcal{L})}\mathbf{w}_n^t,
    \end{equation}
    where $\mathcal{L} = \{l_n \mid n = 1, \ldots, N\}$ denotes the cached dataset length at the $n$ local model. Considering \emph{FedAvg}'s sensitivity to corrupted weights, \gls{raf} is designed by exploring alternatives to the mean \cite{campos2024fedrdf}. Common approaches for robust aggregation are exemplified here: 
    \begin{itemize}
     \item \emph{Trimmed Mean}: In this approach, a parameter $a<\frac{N}{2}$ is chosen. Subsequently, the server eliminates the lowest $a$ values and the highest $a$ values and then computed from the remaining $N-2a$ values, $L$ is then modified as $\mathcal{L}_a = \{l_n \mid 1\leq n\leq N-2a\}$, with 
     \begin{equation}
    \mathbf{w}_g^t = \sum\limits^{N-2a}_{n=1}\frac{l_n}{\sum(\mathcal{L}_{a})}\mathbf{w}_n^t.
    \end{equation}
     \item \emph{FedMedian}: Instead of averaging weights, \emph{FedMedian} utilizes a median computation,
     \begin{equation}
    \mathbf{w}_g^t = \med(\mathbf{w}_1^t,\cdots,\mathbf{w}_N^t).
    \end{equation}
    \end{itemize}
    Through the use of \glspl{raf}, the global model effectively filters out potential malicious model updates. As demonstrated in \cite{FLRobustAggRey}, the \emph{Trimmed Mean} method requires awareness of attacks, more specifically, the count of local models under attack, and its effectiveness varies across different attack modes. On the other hand, the \emph{FedMedian} excels in resilience against various attacks. Despite these strengths, \glspl{raf} generally have a drawback: the potential to discard usefull model updates, which leads to inferior convergence performance compared to \emph{FedAvg}. To address this limitation, \emph{Campos} et al. proposed  a dynamic framework, that alternates between \emph{FedAvg} and \glspl{raf} for a better convergence than simply employing \glspl{raf} \cite{campos2024fedrdf}.


    \section{Threat model}
    \label{sec:threatmodel} 
    In the above described channel estimation scenario, we contemplate the presence of a certain number $J$ of \glspl{ae}. These \glspl{ae} strategically launching attacks on one or several \glspl{sbs}. The objective is to manipulate the global model through the aggregation process. Launching attacks on all \glspl{sbs} demands significant resources. However, with the advent of \gls{uav} as potential attack platforms, the feasibility of attacks on the majority or entirety of \glspl{sbs} can't be ignored. 
    Considering that the model transmission is exclusively between \gls{sbs} and \gls{mbs}, the use of dedicated links can effectively mitigate \gls{mitm} attacks in the \glspl{sbs} level. 
    This attack can be done through the deployment of malicious \glspl{mu} or \gls{mitm} attack agents between the \glspl{mu} and the \glspl{sbs}. We consider the following attack modes:
    \subsubsection{Outdate mode} \glspl{ae} provide random outdated \gls{csi} without any modification. In the case of \gls{mitm} attack, the attack agent is required to store the outdated \gls{csi} and send back to \gls{sbs}. 
    \subsubsection{Collusion mode} \glspl{ae} will provide the same \gls{csi}. This attack mode is designed to manipulate the model, potentially leading to model collapse, as well as to produce a desired outputted \gls{csi}. This attack mode doesn't require any knowledge of \gls{csi} from the \gls{mitm} attack agent. 
    \subsubsection{Reverse mode} \glspl{ae} will provide reversed version of the \gls{csi}, each value of \gls{csi} will be reversed regarding the mean. This attack mode requires the \gls{mitm} attack agent to possess precise knowledge of the \gls{csi} and perform an additional processing step.
    
    Given the channel estimation model is fine-tuned with dataset ${\mathcal{B}}_n^t = \{(\mathbf{x}_\alpha, \mathbf{y}_\alpha) \mid 1\leq \alpha \leq A\}$
    . The object is to minimize the loss $\ell(\mathbf{y}_\alpha,\hat{\mathbf{y}}_\alpha)$, where $\hat{\mathbf{y}}_\alpha$ is the predicted output corresponding to $\mathbf{x}_\alpha$. A common choice for the loss function is the \gls{mse}, defined as:
    \begin{align}\label{eq:MSE}
    \begin{split}
     \mathrm{MSE} &= \frac{1}{A}\sum\limits^A_{\alpha=1}(\hat{\mathbf{y}}_\alpha-\mathbf{y}_\alpha)^2\\    &= \frac{1}{A}\sum\limits^A_{\alpha=1}(\mathbf{w}{\mathbf{x}_\alpha}-\mathbf{y}_\alpha)^2.
    \end{split}
    \end{align}
    And the average gradient with respect to the loss function and weight is given by:  
    \begin{equation}\label{eq:gradientMSE}
    \begin{split}
    \triangledown_{\mathbf{w}} \ell &= -\eta\frac{1}{A}\sum\limits^A_{\alpha=1}\frac{\partial}{\partial \mathbf{w}}(\mathbf{w}{\mathbf{x}_\alpha}-\mathbf{y}_\alpha)^2\\
    &= -2\eta \frac{1}{A}\sum\limits^A_{\alpha=1}\mathbf{x}_{\alpha}(\mathbf{w}\mathbf{x}_{\alpha}-\mathbf{y}_{\alpha}).
    \end{split}
    \end{equation}
    Referring to Eq.~(\ref{eq:gradientMSE}), if the dataset exhibits a large variety, the effectiveness of collusion mode, capable of injecting malicious gradients towards specific $\mathbf{y}$, is likely to weaken.
    In comparison, reversing a subset of $\mathbf{y}_\alpha$ can effectively interfere with, or negate the useful gradient regardless of the dataset variety. The modified subset cached by the \gls{sbs} can be described as  $\dot{\mathcal{B}}_n^t = \{(\dot{\mathbf{x}}_\beta, \dot{\mathbf{y}}_\beta) \mid 1\leq\beta\leq B\}$, where $\dot{y_\beta}$ represents the modified \gls{csi} through different attack modes. And $\bar{\mathcal{B}}_n^t = \{(\bar{\mathbf{x}}_\gamma, \bar{\mathbf{y}}_\gamma) \mid 1\leq\gamma \leq \Gamma\}$ represents the authentic subset. 
    \section{Methodology}
    \subsection{Robust aggregation function: StoMedian}
    The \gls{bme} framework implements Bayesian inference on the server, thereby mitigating potential degradation in model performance \cite{ijcai2023p851}. Essentially, \gls{bme} has the capacity to dynamically aggregate local models. In the scope of \gls{bme}-based \gls{fl}, \emph{chen} et al. firstly employs client parameters to construct a stochastic filter for the aggregation of each federation round. While showcasing superior accuracy compared to \emph{FedAvg}, \emph{Chen}'s approach, \emph{FedBE}, relies on utilizing a diagonal Gaussian distribution for the stochastic filter, with \cite{Chen2020FedBEMB}
    \begin{equation}\label{eq:diagonal1}
    \mu = \sum\limits_n \frac{l_n}{\sum(\mathcal{L})}\mathbf{w}_n,
    \end{equation} 
    \begin{equation}\label{eq:diagonal2}
    \Sigma_{\text{diag}} = \text{diag}\bigg(\sum\limits_n \frac{l_n}{\sum(\mathcal{L})}(\mathbf{w}_n-\mu)^2\bigg). 
    \end{equation}
    Given the sum set of all cached dataset $\mathcal{B}$, $p(\mathbf{w}|\mathcal{B})$ is posterior distribution of $\mathbf{w}$ being learned. The output probability of $\mathbf{w}$ can be denoted as $p(y|x;\mathbf{w})$, and integrating the outputs of all possible model:
    \begin{equation}\label{eq:pBME1}
     p(y|x;\mathcal{B}) = \int p(y|x; \mathbf{w})p(\mathbf{w}|\mathcal{B}) \, d\mathbf{w}.
    \end{equation}
    \gls{bme} with $S$ samples can be described:
    \begin{equation}\label{eq:pBME2}
    p(y|x;\mathcal{B}) \approx \frac{1}{S} \sum_{s=1}^{S} p(y|x; \mathbf{w}^{(s)}),
    \end{equation}
    \begin{equation}\label{eq:pBME3}
    \{\mathbf{w}^{(s)} \sim \mathcal{N}(\mu, \Sigma_{\text{diag}})\}_{s=1}^S.
    \end{equation}

    Building upon this approach, it can be seamlessly integrated into robust aggregation techniques, ensuring system resilience without compromising convergence performance across federation rounds.
    However, it's worth noting that weights in \gls{ml} models are typically normalized to approach $0$, making original approach insensitive when encountering various adversarial attacks. Hence, integrating \gls{bme} into our proposed aggregation function, \emph{StoMedian}, introduces two significant enhancements geared towards improving sensitivity to adversarial attacks:
    \begin{enumerate*}[label=\emph{\roman*)}]
    \item The stochastic filter was constructed using the logarithm of weights to ensure high sensitivity.
    \item Rather than employing $\mu$, we utilized the median of weights to mitigate the potential impact of invaded local models. 
    \end{enumerate*}
    The aggregation function is then presented in Alg.~\ref{alg:stoM}.
    \begin{figure}
		\centering
		\includegraphics[width=0.85\linewidth]{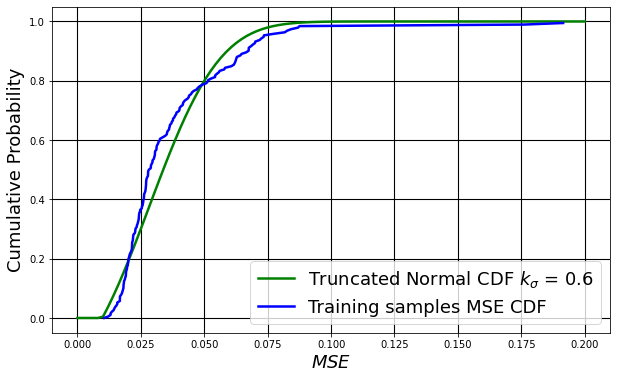}
		\caption{CDF of actual samples and truncated Gaussian distribution}
		\label{fig:DiaUAV}
	\end{figure}
    \begin{algorithm}[!htbp]
    \caption{\small{\emph{StoMedian}}}
    \label{alg:stoM}
    \scriptsize
    \DontPrintSemicolon
    Input: the length of cached dataset $\mathcal{L}$ with respect to the \gls{sbs} ; a small value $\epsilon$ being added to the weights to avoid logarithm of zero. 

    Output: global model weight of current model update: $\mathbf{w}_g$.
    
    \SetKwProg{Fn}{Function}{ :}{end}
    \Fn{\emph{StoMedian}} { 
        get all weights $\{w_{n,j} \mid 1\leq n\leq N; 1\leq j\leq J\}$ from agents;\\
        initialize $ \{p_{n,j} \mid n = 1, \ldots, N; j = 1, \ldots, J\}$ ;\\
        $\mathbf{w}_j = \{w_{n,j} \mid1\leq n\leq N\}$; $\mathbf{p}_j = \{p_{n,j} \mid 1\leq n\leq N\}$\\
        $\mathbf{w}_n = \{w_{n,j} \mid1\leq j\leq J\}$; $\mathbf{p}_n = \{p_{n,j} \mid1\leq j\leq J\}$\\
        \For {each $w_{n,j}$}{\uIf {$w_{n,j}>0$}{$w_{n,j}^{\circ} = -\log(w_{n,j}+\epsilon)$} \ElseIf{$w_{n,j}\leq0$}{$w_{n,j}^{\circ} = \log|w_{n,j}-\epsilon|$}}
       \For {$j = 1:J$ }
         {$\mathbf{w}_j^{\circ} = \{w_{n,j}^{\circ} \mid n = 1, \ldots, N\}$;\\
         $\mu_j = \med(\mathbf{w}_j^{\circ}$);\\
         $\sigma_j = \std(\mathbf{w}_j^{\circ})$;\\
        $\sigma_j = \max(\sigma_j,\epsilon)$;\\
        \For {$n = 1:N$}{$p_{n,j} = \frac{1}{\sigma_j\sqrt{2\pi}}e^{-\frac{(w_{n,j}^{\circ} - \mu_{j})^2}{2\sigma_j^2}}$ }
        $\mathbf{p}_{j} = \frac{\mathbf{p}_{j}\odot \mathcal{L}}{\sum(\mathbf{p}_{j}\odot \mathcal{L})}$ }
        $\mathbf{w}_g = \sum\limits_n \mathbf{p}_n\odot\mathbf{w}_n $

    }
    \end{algorithm}
    
     \begin{algorithm}[!htbp]
    \caption{\Gls{llpf}}
    \label{alg:llpf}
    \scriptsize
    \DontPrintSemicolon
    Input: cached dataset $\mathcal{B}_n$ with respect to the \gls{sbs}; variance modifier index $k_{\sigma}$; determination threshold $\theta$\\
    Output: modified $\mathcal{B}_n$.\\
    \SetKwProg{Fn}{Function}{ :}{end}
    \Fn{\emph{LLPF}}{ 
        initialize $\mathcal{B}_t$ and $\mathcal{B}_{ut}$ to put trustworthy or untrustworthy data; \\
        get $\mathcal{E} = \{\ell_{\alpha} \mid \alpha = 1, \ldots, A\}$ using global model; \\
        $\mu = \med(\mathcal{E})$\\
        Construct $\Phi_{tg}$ with $\mu,k_{\sigma}$\\
        \For {$\alpha = 1:A$ }
        { \uIf {$\Phi_{tg}(\ell_{\alpha})>\theta$}{include $(\mathbf{x}_{\alpha},\mathbf{y}_{\alpha})$ in $\mathcal{B}_t$}
        \ElseIf{$\Phi_{tg}(\ell_{\alpha})\leq\theta$}{include $(\mathbf{x}_{\alpha},\mathbf{y}_{\alpha})$ in $\mathcal{B}_{ut}$}}
        $L_1 = |\mathcal{B}_{ut}|$; $L_2 = |\mathcal{B}_{t}|$ \\
        \For {each $(\mathbf{x}_{l_1},\mathbf{y}_{l_1}) \in \mathcal{B}_{ut}$}{ $r \sim \mathcal{U}(1,L_2)$\\
        replace  $(\mathbf{x}_{l_1},\mathbf{y}_{l_1})$ with $(\mathbf{x}_{r},\mathbf{y}_{r}) \in \mathcal{B}_{t}$ }
        $\mathcal{B}_n = \mathcal{B}_{ut}\cup \mathcal{B}_t$}
    \end{algorithm}
    \subsection{Local loss pre-filtering}
   
    In response to potential attacks employing a widely launched strategy, where most or all local models may be impacted, a common approach is to filter the anomalies before training. Given that the global model has been pre-trained and undergoes fine-tuning during each federation round for enhanced performance, it can be utilized to identify anomalies. In a departure from the conventional use of \gls{lid} for data discrimination, we propose an alternative approach to evaluate the reliability of data originating from each \gls{mu}. The loss of each sample, $\mathcal{E}= \{\ell_{\alpha} \mid 1\leq\alpha\leq A\}$, where $\ell_{\alpha} = (\hat{\mathbf{y}}_\alpha-\mathbf{y}_\alpha)^2$, is computed. The distribution of $\ell_{\alpha}$ heavily relies on the current model and is not mathematically traceable. Therefore, we evaluate it numerically with $300$ randomly generated samples on a pre-trained model.
    Further details on the training setting, model specifications, and data description are provided in Section \ref{sec:evaluate}. As depicted in Fig.~\ref{fig:DiaUAV}, the \gls{cdf} of $\ell_{\alpha}$ can be accessed with \gls{cdf} of a truncated Gaussian distribution with lower and upper bounds of $[0,+\infty)$, with
    \begin{equation}\label{eq:CDFguassian}
    \Phi_{tg}(x) = \frac{1}{2} + \frac{1}{2} \cdot \text{erf}\left(\frac{k_{\sigma}\mu}{\sqrt{2}} \left( x - \mu \right)\right),
    \end{equation}
    \begin{equation}\label{eq:CDFguassian2}
    \mu = \sum\limits_{\alpha = 1}^{A}\ell_{\alpha}.
    \end{equation}
    where $k_{\sigma}= 0.6$, which is determined empirically. Similar to the approach introduced in \cite{han2024secure}, by assigning a threshold $\theta$ to $\Phi_{tg}$, it is capable of detecting anomalies. Further details of the algorithm can be found in Alg.~\ref{alg:llpf}.

    \section{Evaluation}\label{sec:evaluate}
    \subsection{Model and datasets}
    In this work, we employ a simple \gls{cnn} for channel estimation, comprising:
    \begin{enumerate*}[label=\emph{\roman*)}]
    \item an input layer ($5\times5$ kernel size, $24$ filters) with SeLU activation,
    \item a hidden layer ($5\times5$ kernel size, $8$ filters) with Softplus activation,
    \item an output layer ($5\times5$ kernel size, $1$ filter) with SeLU activation.
    \end{enumerate*}
    This \gls{cnn} processes input data of size $612\times14\times1$ and has demonstrated effective channel estimation performance, as evidenced by \cite{defCatakCE2022}.
    
    The datasets used for training channel estimation models were generated using a reference example titled "Deep Learning Data Synthesis for 5G Channel Estimation," accessible within the MATLAB 5G Toolbox, with specific settings considered. Given that pilot signals and \gls{csi} are typically represented as complex numbers, we adopted a preprocessing step. Specifically, we divided the complex data into its real and imaginary components to make it compatible with \gls{cnn}. This approach aligns with prior work.
    \begin{figure*}[!htbp]
    \centering
    \begin{subfigure}[b]{0.434\textwidth}
        \centering
        \includegraphics[width=\linewidth]{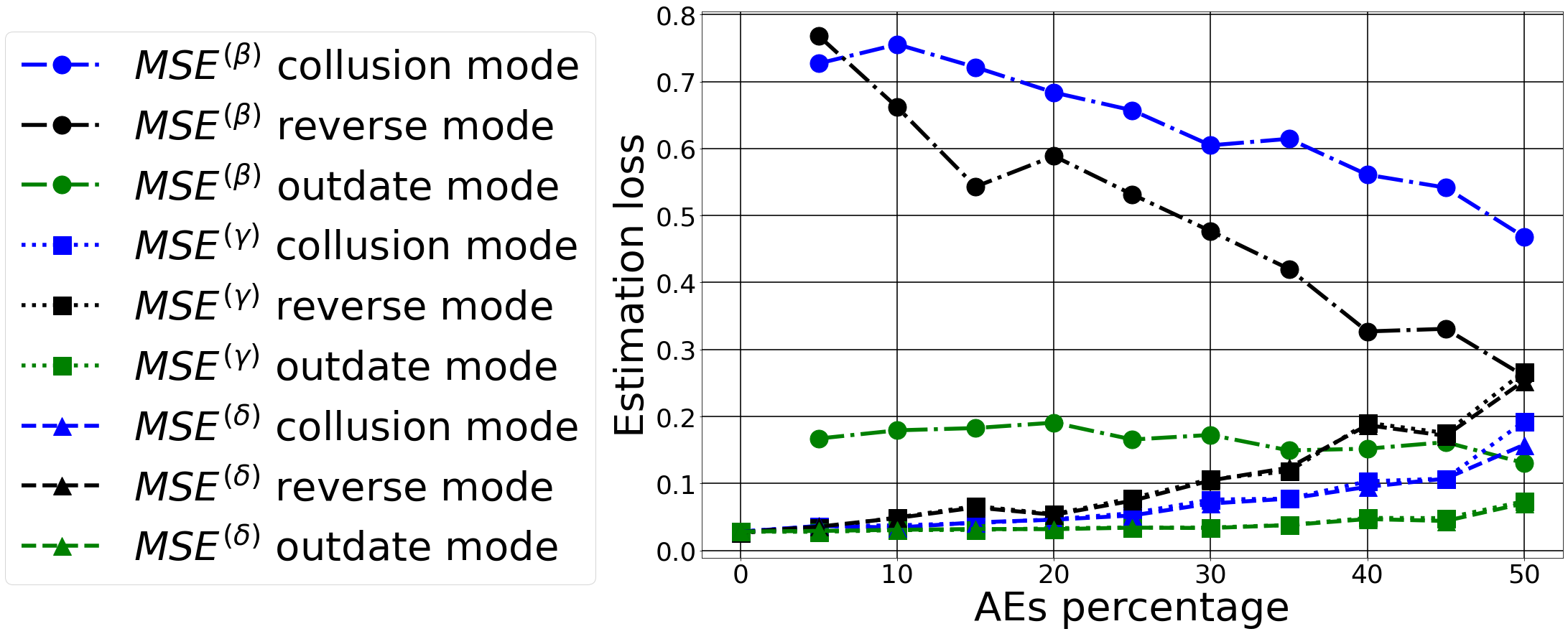}
        \caption{Attack evaluation of \gls{cl}}
        \label{fig:cl}
    \end{subfigure}
    \begin{subfigure}[b]{0.27\textwidth}
        \centering
        \includegraphics[width=\linewidth]{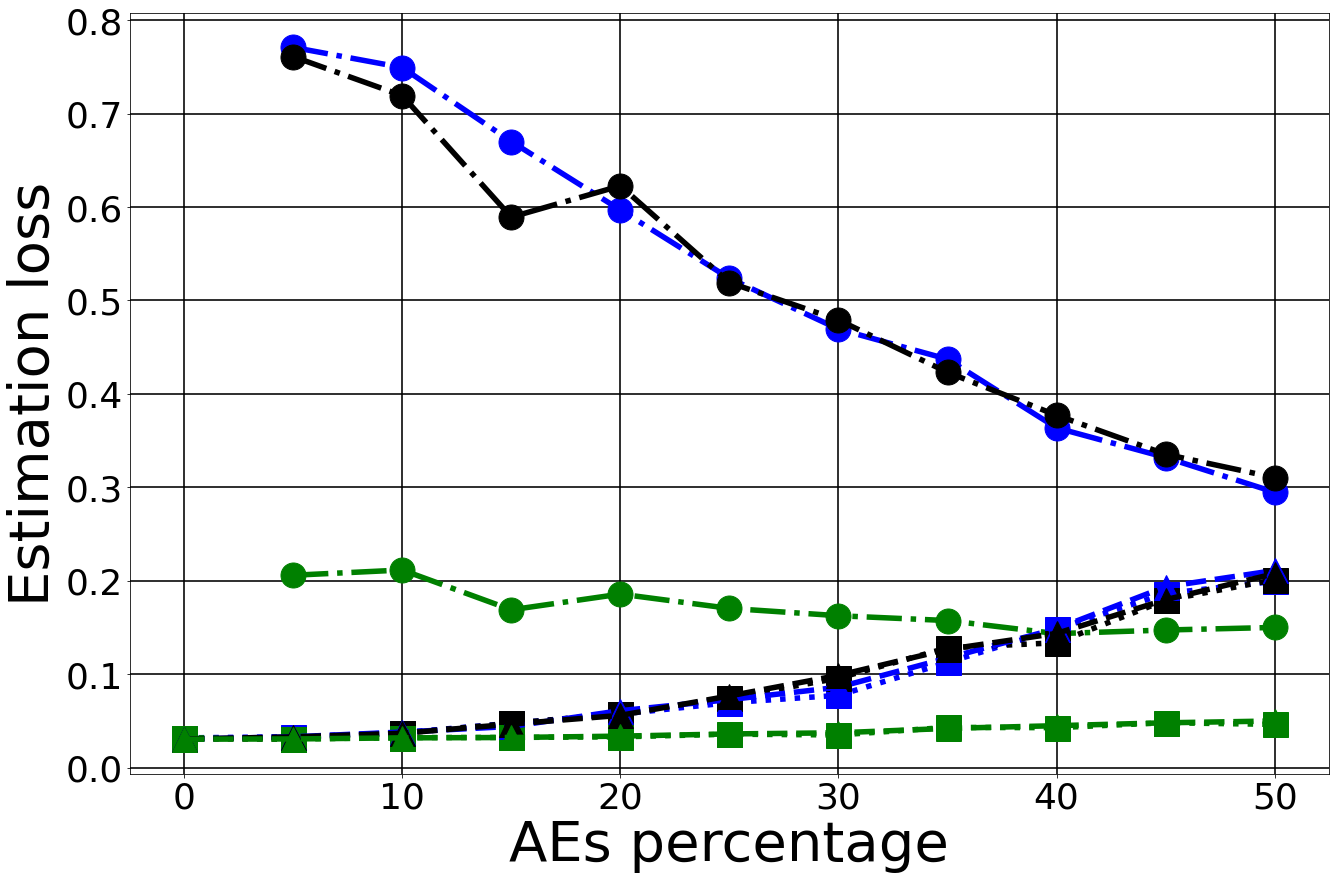}
        \caption{\gls{fl} during widely launched attacks}
        \label{fig:randomdep}
    \end{subfigure}
    \begin{subfigure}[b]{0.27\textwidth}
        \centering
        \includegraphics[width=\linewidth]{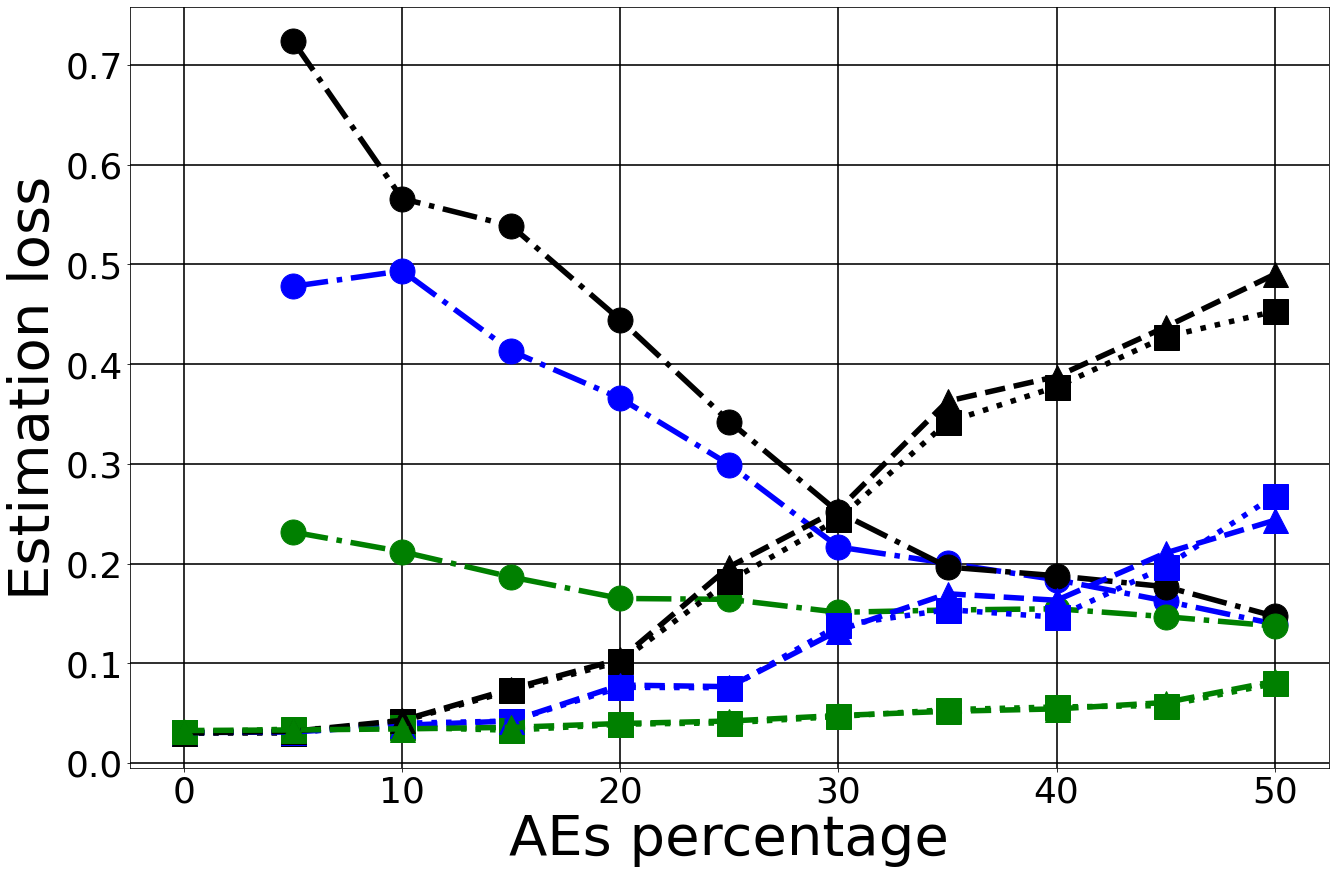}
        \caption{\gls{fl} during targeted attacks}
        \label{fig:coordinateddep}
    \end{subfigure}    
    \caption{Threat evaluation}
    \label{fig:threat-eval}
\end{figure*}

    \subsection{Threat evaluation}\label{subsec:Tevaluation}
    The global model was initially pre-trained with a small dataset including $200$ samples. Subsequently, local models were fine-tuned with $\mathcal{B}_n^t$, $ \mathcal{B}_n^t = \dot{\mathcal{B}}_n^t \cup \bar{\mathcal{B}}_n^t $. A validation dataset $\mathcal{B}^\circ = \{(\mathbf{x}_\delta^\circ, \mathbf{y}_\delta^\circ) \mid 1\leq \delta \leq \Delta\} $, was employed to assess the channel estimation performance, with these data  never being part of the training.   

    To demonstrate the susceptibility of \gls{fl}, we simulated our \glspl{sbs} under data poisoning attacks, alongside a \gls{cl} scenario as comparison. The standard \emph{FedAvg} aggregation function was used. We define $\mathrm{MSE}^{(\gamma)} = \mathrm{MSE}(\bar{\mathbf{y}}_\gamma, \hat{\mathbf{y}}_\gamma)$ as an indicator of the model's performance on seen data, $\mathrm{MSE}^{(\delta)} = \mathrm{MSE}(\mathbf{y}^\circ_\delta, \hat{\mathbf{y}}_\delta)$ as an indicator of its performance on unseen data, and $\mathrm{MSE}^{(\beta)} = \mathrm{MSE}(\dot{\mathbf{y}}_\beta, \hat{\mathbf{y}}_\beta)$ as a measure of the efficacy of adversarial manipulations on the model.
    
    We considered a minimum cached dataset length required for training. In cases where the \gls{sbs} lacks sufficient data, a subset of pre-training data is appended to meet the minimum requirement. \glspl{ae} can be deployed either in a widespread strategy or to target a specific \gls{sbs}. The simulation setup is described in Tab~\ref{tab:setup1}. 
    \begin{table}[!htbp]
		\centering
        \scriptsize
		\caption{Simulation setup 1}
		\label{tab:setup1}
		\begin{tabular}{>{}m{0.2cm} | m{1.4cm} l m{3.5cm}}
			\toprule[2px]
			&\textbf{Parameter}&\textbf{Value}&\textbf{Remark}\\
			\midrule[1px]
            &$U$&$1000$& Number of \glspl{mu}\\    
			  &$N$&$10$& Number of \glspl{sbs}\\
			\multirow{-2}{*}{\rotatebox{90}{\textbf{System}}} &$M_{b}$&$1$& Number of \gls{mbs}\\
            &$K$&$10$& Number of model update rounds\\
            &$r_{a}$&$0\%\sim50\%$& Ratio of \glspl{ae}\\

			    \midrule[1px]
		
			&$I_{min}$ & $200$& Minimum length of cached datasets required for training\\
            &$\sum(L)$ & $2000$& Length of all cached datasets of one federation round\\
            &$l_n$ & $l_n\sim\mathcal{U}(170,230)$& Length of a local cached dataset (without targeted attacks)\\
            &$\Delta$ & $200$& Length of validation dataset\\
			&$l_r$ & $0.001$& Learning rate for adam optimizer\\
	          &$m$ & $0.9$& Momentum for adam optimizer\\
			&$E$ & $100$ & Epochs of the training\\
            \multirow{-10.5}{*}{\rotatebox{90}{\textbf{Training}}} &$N_{batch}$ & $64$ & Batch size\\
            \bottomrule[2px]
		\end{tabular}
	\end{table}
    
    As analyzed in Section~\ref{sec:threatmodel}, the reverse mode can nullify useful gradients learned from authentic data, a phenomenon whose effectiveness is confirmed by our simulation results in Figs.~\ref{fig:cl}{--}\ref{fig:coordinateddep}. Conversely, the outdate mode has minimal impact, only slightly degrading model accuracy, due to the strong correlation between outdated \gls{csi} and actual \gls{csi}. Combining Fig.~\ref{fig:cl} and Fig.~\ref{fig:randomdep}, \gls{fl} shows significant performance degradation under collusion attacks compared to \gls{cl}. This is attributed to the fact that \gls{fl} trains local models on a small dataset, characterized by a limited diversity. 
    
    Fig.~\ref{fig:coordinateddep} demonstrates the vulnerability of \gls{fl} to targeted data poisoning attacks on specific clients. For reverse attack, at $30\%$, intersections between $\mathrm{MSE}^{(\beta)}$ and either $\mathrm{MSE}^{(\gamma)}$ or $\mathrm{MSE}^{(\delta)}$ become evident. This signifies the successful evasion of the attacker, making the attacks undetectable by the \gls{ml} model itself. Meanwhile, the intersections of 
    collusion attack appear at $50\%$. Both attack modes are significantly boosted because \emph{FedAvg} employs a weighted approach. This results in the victim local model, surrounded by \glspl{ae}, being assigned a larger weight, further degrading model performance during aggregation.
    However, the efficacy of such attack strategies can be efficiently mitigated by employing \glspl{raf}.
    \subsection{Evaluation of proposed methodology: StoMedian}
     \begin{figure}
    \centering
    \begin{subfigure}[b]{\linewidth}
        \centering
        \includegraphics[width=0.85\linewidth]{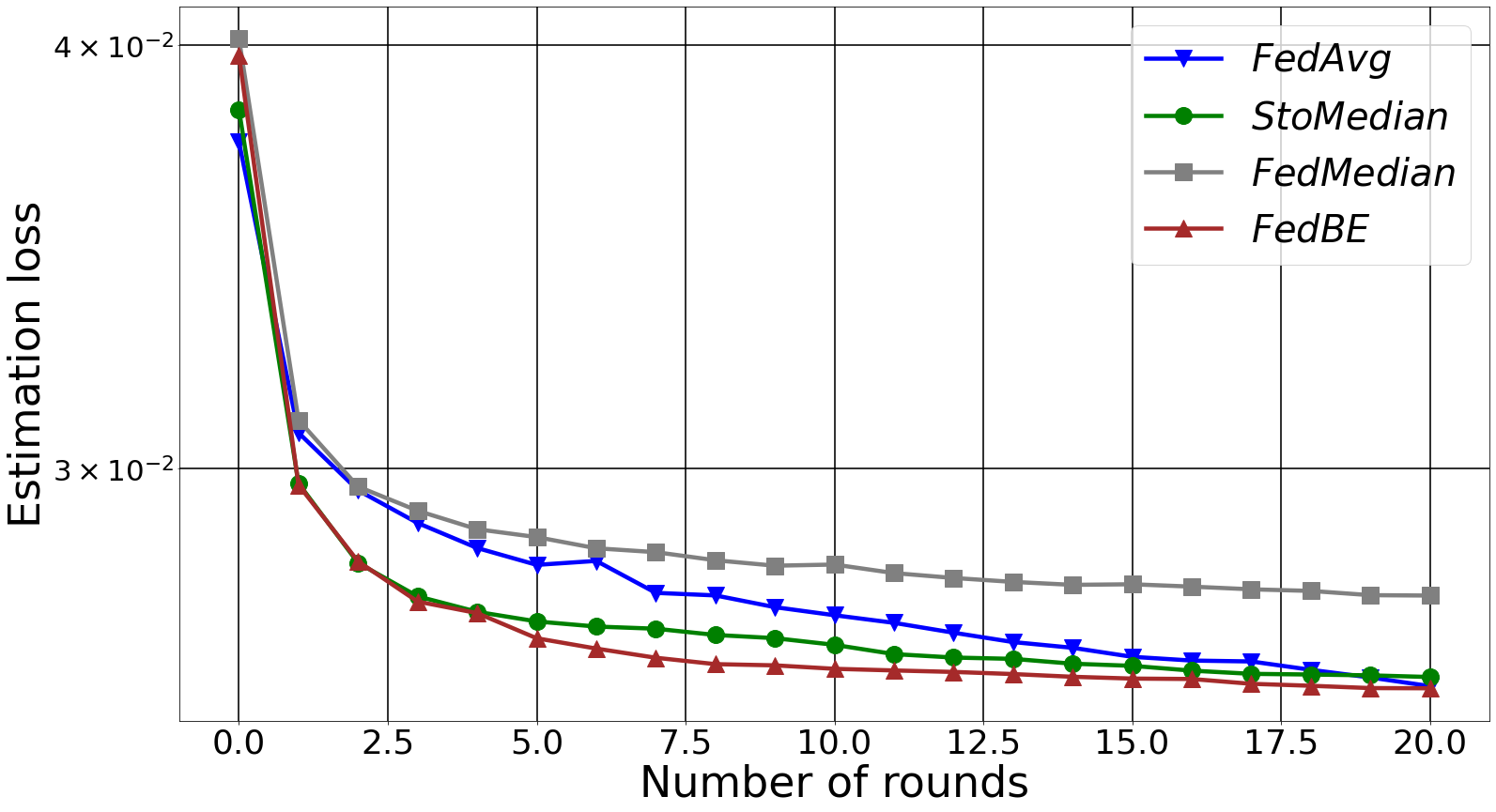}
        \caption{Convergence results without attacks}\label{fig:no_attack}
    \end{subfigure}\\[1ex] 
    \begin{subfigure}[b]{\linewidth}
        \centering
        \includegraphics[width=0.88\linewidth]{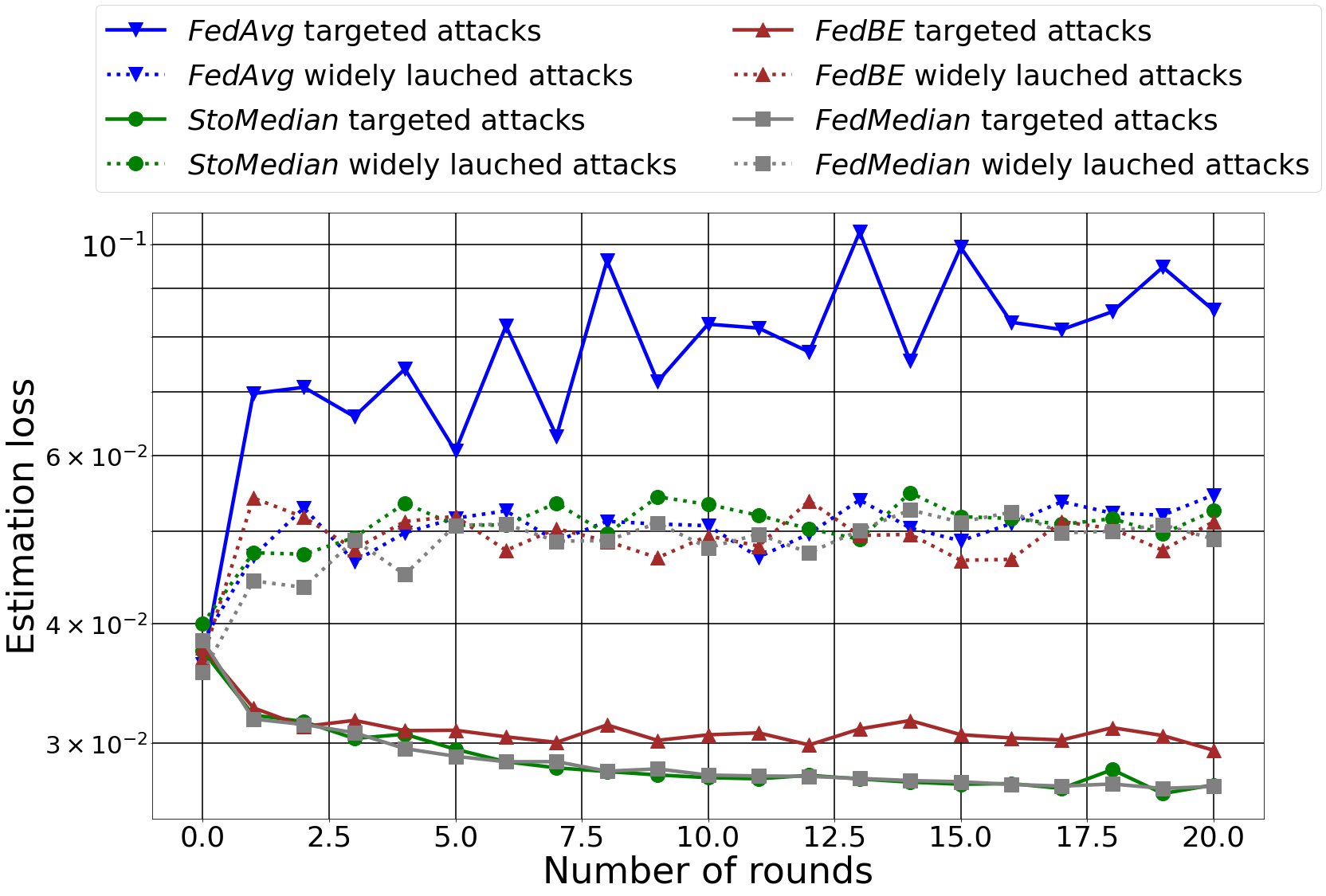}
        \caption{Convergence results while under attacks}\label{fig:attack}
    \end{subfigure}
    \caption{Convergence test of \gls{raf}}
   \end{figure}
    While \glspl{ae} are being deployed to target a specific \gls{sbs}, which is a cost-efficient and most impactful attack strategy from attacker side. To demonstrate the effectiveness of \glspl{raf}, we conducted simulations to illustrate the convergence of $\mathrm{MSE}^{(\delta)}$ across $20$ federation rounds for four different aggregation functions. The simulation setup is same as listed in Tab~\ref{tab:setup1}, with only reverse attack being considered and $r_a = 20\%$.  
    
    Firstly, we examine a scenario devoid of any attacks, depicted in Fig.\ref{fig:no_attack}. Unsurprisingly, \emph{FedMedian} demonstrates the poorest convergence performance, as we elucidated in Sec.\ref{subsec:rma}. \emph{StoMedian} and \emph{FedBE} demonstrated significantly superior performance compared to \emph{FedAvg}, primarily due to its effectiveness in preventing model drift in the aggregation. As the overall performance improved over federation rounds, the performance gap became less evident, with all three aggregation function reaching a comparable convergence level. Meanwhile, \emph{FedBE} performed slightly better than \emph{StoMedian}.  Nevertheless, the performance of \emph{StoMedian} is contingent upon the distribution of underlying local datasets. In the worst-case scenario where local data are highly non-identically distributed, which is unlikely in our setting where \glspl{sbs} have confined coverage and are not likely to exhibit diverse geographical deployments. \emph{StoMedian} may filter out most local weights, resulting in convergence levels similar to that of \emph{FedMedian}. In such a scenario, the performance degradation of \emph{FedAvg} is inevitable. 
    
    The results of convergence analysis of aggregation functions under attacks are depicted in Fig.\ref{fig:attack}. \emph{FedAvg} exhibited no defense against attacks, and \emph{FedBE} demonstrated limited defense but far from resilient. The performance gap between \emph{FedBE} and \emph{FedMedian} is larger than that of Fig.\ref{fig:no_attack}, due to the scaling which may not be immediately evident.  In contrast, \emph{FedMedian} and \emph{StoMedian} demonstrated resilient and comparable performance against coordinated strategy. On the other hand, all aggregation functions failed to mitigate widely launched attacks effectively. 
     \subsection{Evaluation of proposed methodology: LLPF}
     Having evaluated the effectiveness of \emph{StoMedian}, we now proceed to assess the efficacy of \gls{llpf}. The simulation setup remains consistent with the parameters outlined in Table~\ref{tab:setup1}, with \emph{StoMedian} being applied. We examine three attack modes, with $r_a$ set at $20\%$. 

\begin{figure}
    \centering
    \begin{subfigure}[b]{\linewidth}
        \centering
        \includegraphics[width=0.90\linewidth]{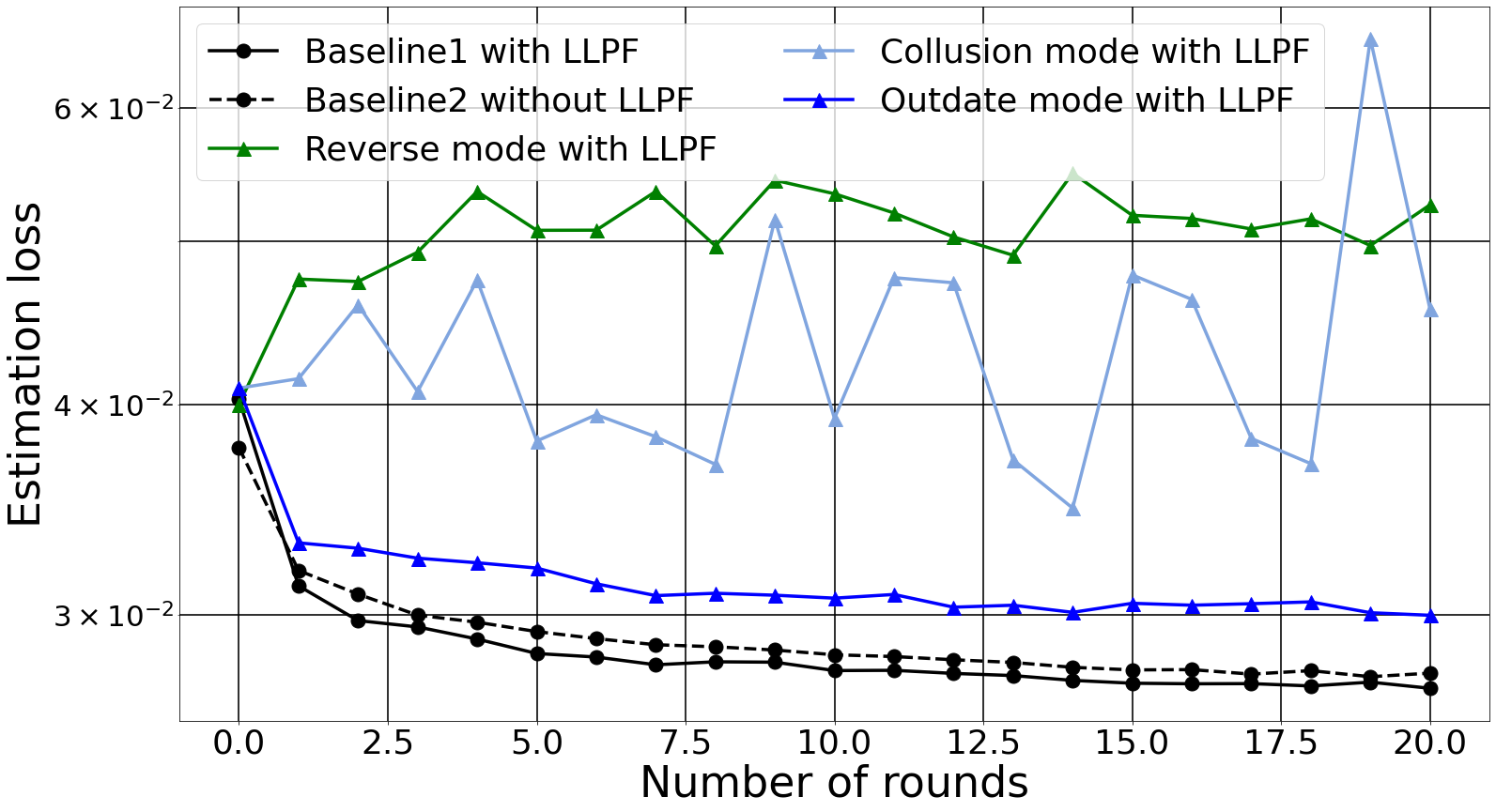}
        \caption{Convergence results without \gls{llpf} under 3 attack modes during widely launched attacks}\label{fig:WithoutLLPF}
    \end{subfigure}\\[1ex] 
    \begin{subfigure}[b]{\linewidth}
        \centering
        \includegraphics[width=0.90\linewidth]{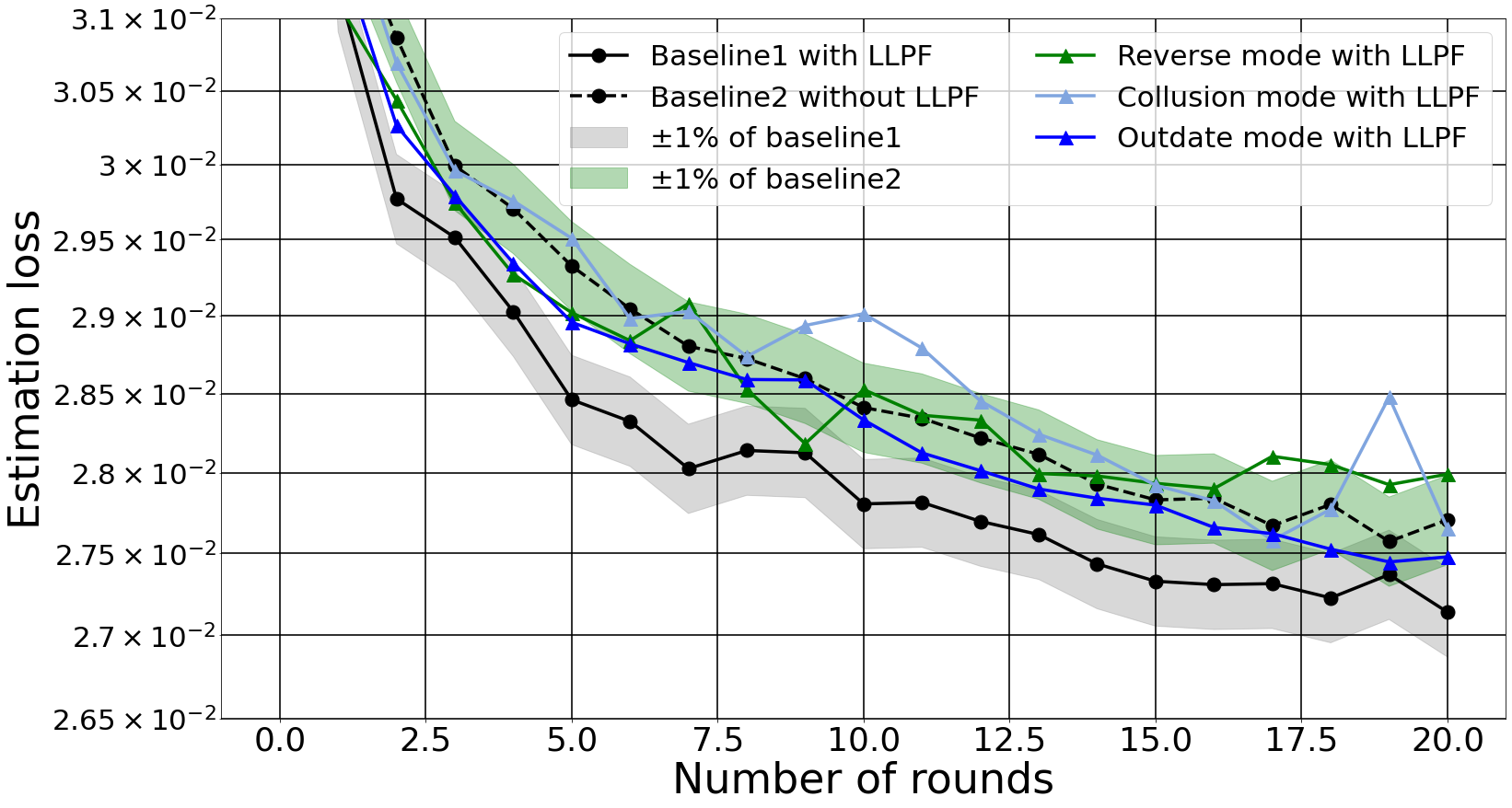}
        \caption{Convergence results with \gls{llpf} ($\theta= 0.95$ $k_{\sigma} = 0.6$) under 3 attack modes during widely launched attacks}\label{fig:WithLLPF}
    \end{subfigure}
    \caption{Convergence test of \gls{llpf}}\label{fig:LLPF}
   \end{figure}

     We compared the simulation results with two baselines, obtained by excluding $20\%$ of the authentic data. The results depicted in Fig.\ref{fig:LLPF} indicate that employing \gls{llpf} leads to a slight improvement in convergence performance. When comparing the reverse mode and collusion mode in Fig.~\ref{fig:WithoutLLPF}, the attack performance of collusion mode exhibits high fluctuations. This observation aligns with our analysis in Sec.\ref{sec:threatmodel}, confirming that the performance of collusion attacks strongly depends on the variety of data. 

     In Fig. \ref{fig:WithLLPF}, \gls{llpf} exhibited robust performance against all attack modes, with performance levels roughly converging to $\pm1$ of \emph{baseline2}. However, considering a certain level of miss-detection is inevitable, this prevents the convergence performance with \gls{llpf} from reaching the level of \emph{baseline1}. In summary, \gls{llpf} effectively mitigate widely launched attacks as well as improve the overall performance. 
     
    \section{Conclusion and outlook}
    This work has demonstrated the vulnerability of \gls{fl} when applied in wireless communication, illustrated through a channel estimation scenario. Various adversarial attacks and strategies were devised to exploit this vulnerability. Subsequently, \emph{StoMedian} was introduced to enhance resilience during the aggregation process, along with \gls{llpf} tailored for incremental training for channel estimation. Both \emph{StoMedian} and \gls{llpf} have been validated through simulation. 

    However, the performance of the mentioned methods may be impacted when the cached data exhibit highly non-identical distributions, a factor not addressed in our paper. Particularly for \gls{llpf}, in scenarios with non-identical or highly varied data, it may filter out new features for the model to learn. To address this, a soft determination process based on the cumulative trustworthiness of \glspl{mu}, as exemplified in \cite{softdeter2023fang}, could be applied. By integrating the current determination with the \gls{mu}'s historical behaviors, a smart determination process could further enhance model convergence.
    \section*{Acknowledgment}
	This work is supported partly by the German Federal Ministry of Education and Research within the project Open6GHub (16KISK003K/16KISK004), partly by the European Commission within the Horizon Europe project Hexa-X-II (101095759). B. Han (bin.han@rptu.de) is the corresponding author.
    
\bibliographystyle{IEEEtran}
\bibliography{references}

\end{document}